# The Capability of Large Language Models to Measure Psychiatric Functioning

Isaac R. Galatzer-Levy[1], Daniel McDuff[1], Vivek Natarajan[1], Alan Karthikesalingam[1], Matteo Malgaroli[2]

[1]Google Research
[2]NYU Grossman School of Medicine

**Abstract**

The current work investigates the capability of Large language models (LLMs) that are explicitly trained on large corpuses of medical knowledge (Med-PaLM 2) to predict psychiatric functioning from patient interviews and clinical descriptions without being trained to do so. To assess this, $n = 145$ depression and $n = 115$ PTSD assessments and $n = 46$ clinical case studies across high prevalence/high comorbidity disorders (Depressive, Anxiety, Psychotic, trauma and stress, Addictive disorders) were analyzed using prompts to extract estimated clinical scores and diagnoses. Results demonstrate that Med-PaLM 2 is capable of assessing psychiatric functioning across a range of psychiatric conditions with the strongest performance being the prediction of depression scores based on standardized assessments (Accuracy range= 0.80 - 0.84) which were statistically indistinguishable from human clinical raters $t(1,144) = 1.20; p = 0.23$. Results show the potential for general clinical language models to flexibly predict psychiatric risk based on free descriptions of functioning from both patients and clinicians.

**Main text**

Assessment of psychiatric functioning represents a common task across verticals of medicine. Primary care settings are the most common first point of contact for treatment or triage of common psychiatric disorders including depression, anxiety, post traumatic stress, psychosis, and addiction.[1] As such, it is a public health priority to scale up the assessment of common psychiatric risk and illness in primary care settings.[2,3] Assessments rely on either verbal self-report or structured screening instruments, often administered by non-experts with limited experience in determining risk or adjudicating between disorders.[4] The lack of standardization and automation of clinical information scoring and assessment represents a unique limitation of psychiatry compared to other areas of medicine. This limitation can be attributed to the nature of psychiatric assessment which results in linguistic descriptions rather than biological values that can be mathematically parsed.[5] Large language models (LLMs), which utilize advances in neural network architecture that are trained on large text datasets to flexibly interpret and respond to natural language, have demonstrated emergent learning capabilities whereby they can solve natural language problems, such as translation, that they were not explicitly trained to solve.

Large scale language models (LLMs) are trained using relatively simple pre-text tasks, involving predicting preceding, intermediate or subsequent words or sentences that are hidden from the input [6]. These models have been shown to capture complex knowledge and concepts, due to scaling the neural architectures and the data used to train them.[6–8] The models can produce text that is indistinguishable from that written by humans,[9] match human-level performance on multiple reading-comprehension benchmarks, and can achieve passing grades on medical and law bar exams.[10,11] The capability to learn patterns in data without providing training examples, known as self-supervised learning, comes from training on large numbers of parameters and data sources to learn general rules and relationships that can be applied to answer specific questions. Just as LLMs trained on large corpuses of language can learn translation without training by understanding general rules of language, so may LLMs trained on general medical knowledge be capable of understanding specific linguistically based rules of clinical assessment.

A particularly fruitful area to apply LLMs within medicine may be psychiatric assessment. Screening and diagnosis of psychiatric risk is linguistically assessed and communicated. As a result, machine learning models that utilize example data such as clinical interviews to train a model, known as supervised learning, have demonstrated strong results in classifying disorders including depression, PTSD, and psychosis.[12] However, because these models are trained on relatively narrow examples, they are typically not flexible enough to be applied to data that is different then the training examples. Owing to the large number of parameters and underlying training data, LLMs likely have latent knowledge of psychiatric language, assessment, symptoms and diagnosis. This knowledge is inconsistent and likely requires further training on focused medical knowledge (e.g. journals, case studies). Such data sources can provide diverse sources of psychiatric knowledge as screening and assessment of common psychiatric disorders commonly occur in non-psychiatric settings.[13] Further, as LLM technology is applied to medical applications, there is additional regulatory and ethical obligation to ensure that AI-driven assessments are built on sound data sources.[14] The current work tests the capabilities of LLMs to generate predictions of psychiatric symptom severity and diagnoses.

*Depression and PTSD clinical interview assessments*
We implement our experiments on a transformer architecture (PaLM 2 [8]), with medical domain fine-tuning called Med-PaLM 2.[11] For the purposes of this analysis the large (L) model was used. Pretrained on a massive text corpus ([10,11]) comprising hundreds of billions of tokens, the model has been exposed to a diverse set of natural language use cases drawn from various sources (for full description of Med-PaLM 2 development (see [11]) .First, to assess Med-PaLM 2's accuracy to measure and screen for depression and post traumatic stress disorder (PTSD), we utilized research grade clinical interview transcripts (for full description see *methods*). Transcripts were entered as inputs to Med-PaLM following a standardized prompt structure designed to 1) focus model attention on knowledge of the utilized PTSD and depression rating scales (PCL-C; PHQ 8), 2) estimate scores on both scales; 3) produce a confidence estimate; 3) provide descriptive reasoning for the selected score (see *methods* for full description).

Med-PaLM 2 produced estimated scores for the PHQ-8 and PCL-C respectively. Estimated scores with no training ($\mu$ = 8.50 ; SD = 9.02) were not statistically different from human raters ($\mu$ = 7.94 ; SD = 5.36) for depression ratings [PHQ-8: $t(1,144)$ = 1.20; $p$ = 0.23] but were significantly different for PTSD ratings [PCL-C; $t(1,114)$ = 2.02; $p<.01$] with humans ($\mu$ = 27.77 ; SD = 11.53) scoring subjects significantly lower than Med-PaLM 2 ($\mu$ = 36.51 ; SD = 12.55). Results of the prediction of real number scores demonstrated that Med-PaLM 2 predicted participant scores at a low error rate and identified caseness of depression and PTSD at a good accuracy. Analyses further revealed that while Med-PaLM 2's classified depression at both high sensitivity and specificity (sensitivity = 0.75; specificity = 0.82), Med-PaLM 2's performance classifying PTSD demonstrated strong specificity (0.98, it demonstrated low sensitivity (0.30; See Table 1 for full results)

In comparison to results of human raters in the published literature, Med-PaLM 2 performance in assessing depression, as measured by sensitivity and specificity is consistent with performance of human raters pooled across studies through meta-analysis (Pooled sensitivity = 0.84 (range = .70- .94) ; Pooled specificity = 0.81 (range = 0.69 - 0.82 ). Further Med-PaLM 2 demonstrated consistent scores on Cohen's Kappa, a measure of agreement between raters, with human raters when compared to estimates in the literature between two human raters Med-PaLM 2 Kappa = .55; Published Kappa ranges = 0.35 - 0.76).[18] The PCL has similarly demonstrated a range across studies (sensitivity range = 0.20- 1.00; specificity range = 0.71 - 0.99) with variability attributed to study bias and a lack of consensus on the appropriate score for a clinical cut-off.[19] In this instance, Med-PaLM 2 demonstrated high specificity but moderate to low sensitivity and a Kappa between the LLM and the human rater of 0.33 indicating fair agreement.[20] Sensitivity, specificity, and the Kappa between human and LLM ratings are reported in Table 1. Results further demonstrated improvements when weighting regression results by the model confidence score and when using a cut-off threshold =>-0.20 (See supplemental Figure 1).

Table 1: Results of Med-PaLM 2 prediction of PCL-C and PHQ-8 scores and clinical cutoffs.

|  | Med-PaLM 2 PCL-C | Med-PaLM 2 PHQ-8 |
| --- | --- | --- |
| Accuracy | 0.74 | 0.80 |
| F1 Score | 0.64 | 0.77 |
| Precision | 0.88 | 0.65 |
| Sensitivity | 0.30 | 0.75 |
| Specificity | 0.98 | 0.82 |
| MAE | 9.07 | 2.33 |
| RMSE | 11.2 | 3.93 |
| Kappa with Clinical Ratings | 0.33 | 0.55 |
| Pearson r (p-value) | 0.41 (p < 0.01) | 0.55 (p < 0.01) |

To assess Med-PaLM 2's capability to extract and summarize diagnostically relevant information, we compared the frequency of diagnostically descriptive terms and phrases taken from the DSM 5 descriptions of MDD and PTSD. We compared the frequency of MDD and PTSD words in the description of PHQ-8 and PCL-C estimates across cases who were assessed for both. Results demonstrated that Med-PaLM 2 was significantly more likely to use words associated with the correct diagnosis when describing results of the PHQ-8 and PCL-C respectively [$\chi^2(1,146) = 138.12$; $p < .001$; O.R. = 3.88; see Supplemental Table 1; Figure 2].

Figure 2: Frequency of words and phrases associated with Major Depressive Disorder (MDD) and Posttraumatic Stress Disorder (PTSD) associated with MDD and PTSD assessments

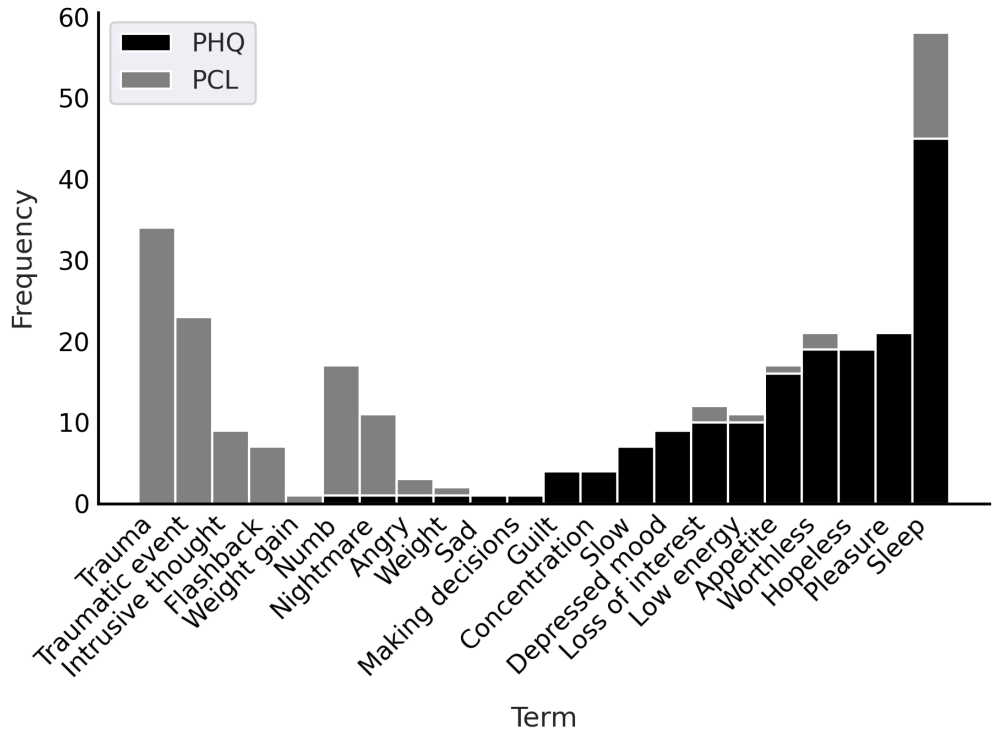

*Note*: Sleep was removed from analyses because sleep abnormalities are diagnostic of both PTSD and MDD.

*Clinical case study assessment*

Next, to assess Med-PaLM 2's capabilities to provide clinical labels across diverse psychiatric disorders that are commonly encountered in primary care setting, we entered individual de identified training case studies with accompanying diagnoses hidden from the following psychiatric categories: depressive disorders (e.g. dysthymia, MDD, premenstrual dysphoric disorder; $n = 12$), anxiety (e.g., specific phobias, Generalized Anxiety Disorder; $n = 6$), posttraumatic (e.g., PTSD, acute stress disorder; $n = 8$), substance and addiction related (e.g., cocaine dependance; gambling disorder; $n = 7$), and psychotic disorders (schizophrenia, schizoaffective disorder; $n = 7$) from American Psychiatric Association (APA) training examples.[21] We assessed the accuracy to correctly label the diagnostic category, label the specific diagnosis within each category compared to others, and label additional diagnoses or diagnostic modifiers (See *methods* for a description of the dataset).

Without providing guidance on diagnoses to choose between, Med-PaLM 2 correctly labeled the diagnostic categories 92.5% (n = 37) of the time and labeled the correct diagnosis 77.5% (n = 31) of the time. The frequency of correct diagnosis within each category was compared using Fisher's Exact Test. Results demonstrated that there were no statistically significant differences between diagnoses, except for Depressive Disorders for which Med-PaLM 2 demonstrated marginally better categorization using Fisher's exact test [Phi (1, 39) = -0.27; $p = .09$; O.R. =

0.32]. All other diagnoses within categories did not approach significance (see Supplemental Table 2). Finally, in 24 case studies, an additional diagnosis or diagnostic modifier was provided. Med-PaLM 2 correctly generated additional diagnoses and modifiers in only 20% (n = 5) of cases. Given the small sample of subjects within each category that had additional diagnoses and modifiers, no statistical tests could be employed for comparison.

In summary, we applied Med-PaLM-2, an LLM trained on general medical knowledge,[10] to predict the results of structured depression and PTSD assessment instruments from de-identified transcripts of standardized research clinical interview databases. Med-PaLM-2 achieved state-of-the-art performance on these tasks, and was additionally able to provide an estimated confidence rating in its assessment, and a written explanation of the rationale for its scoring. Further, when applied to case studies, Med-PaLM-2 demonstrated high accuracy in labeling DSM 5 diagnoses without prior training. Importantly Med-PaLM 2 performed inconsistently in identifying comorbidities and diagnostic modifiers (i.e. MDD with psychotic features) indicating that additional training or prompt tuning may be required to improve models.

In addition to prediction of diagnosis, Med-PaLM 2 provided explanations for the model decisions based on the text. The capability to summarize the reasoning behind modeling decisions is an important advance in machine learning that aims for medical applications. High dimensional models are commonly criticized and limited in their medical applications because they are uninterpretable "black boxes". In the current context where psychiatric diagnoses are heterogeneous in their symptom presentation and treatments and linguistic descriptions can be idiosyncratic among non-specialist clinicians,[22] explanations through summarization are equally relevant to assessment outcomes. To formally test the capability for summarization in PTSD and MDD assessment, we conducted an analysis of the frequency of words associated with each diagnostic category revealed a statistically higher frequency of words and phrases associated with each diagnostic category within Med-PaLM 2's explanation of the model result. This indicates that LLMs both provide diagnoses and provide explanatory summarization at an accuracy that is actionable for clinical screening.

Large language models demonstrate the capability to read and understand common psychiatric constructs without explicitly teaching them to perform this function. Further, analyses of word frequencies show that Med-Palm 2 produces content-specific summarization. As such, Med-PaLM 2 demonstrates the capability to assess psychiatric functioning based on both patient and clinician's descriptions while providing explainable summarization. Results demonstrate that Med-PaLM 2 is better able to predict depression than other psychiatric constructs. Depression ratings from Med-PaLM 2 did not differ statistically from human raters while PTSD scores were shown to be statistically discrepant between human and automated scoring. Further, depressive disorder classification based on case studies was marginally better than other disorder categories. This result likely reflects the overall prevalence of depressive disorders and even the use of the

PHQ in clinical screening which are both much more common than other disorders and screenings. Additional model tuning may be required to improve the accuracy of other assessments. We hypothesize that Med-Palm 2 performed best when assessing MDD because it is the most commonly occurring psychiatric condition and psychiatric comorbidity with physical illness. As a result, models most likely had access to the greatest number and diversity of training examples in the general medical corpus. Further, the current results are demonstrated on relatively small datasets and in limited use cases. As such, the performance of these models can not be generalized and we can not make broad claims about the capability to screen or evaluate psychiatric functioning using LLMs. The disorders assessed in this study are in high prevalence globally while the current work is limited to English only using demographically narrow data sources for testing. Results are not intended to serve as a generalized solution, but rather to demonstrate the capabilities that can be harnessed, developed, and validated to improve the scale and access of psychiatric screening and assessment. Additional data sources from diverse populations and formats are required to generalize and apply these results.

Despite these limitations, the current results demonstrate that LLMs trained on general medical knowledge have emergent capabilities to predict psychiatric functioning without being trained to do so. As such, LLMs are likely to find broad applications to standardize screening and assessment across medical contexts that rely on verbal descriptions from patients and clinicians.

**Methods**

*Datasets*

*Depression and PTSD clinical interview assessments*
The Distress Analysis Interview Corpus Wizard of Oz (DAIC-WOZ) and extended corpus [15] was utilized as a primary data source for analysis. The DAIC-WOZ contains previously de-identified interview transcripts and accompanying expert ratings on the 8 item Patient Health Questionnaire [(PHQ-8) [16]; n=145] and the PTSD Checklist-Civilian version [(PCL-C [17]); n =115]. These data are available to researchers from the original publisher (see [15]) and was obtained through an investigator initiated request. Data sources were deemed to not require IRB oversight because they were public and de-identified. Based on the clinical cut-off of 10 on the PHQ-8, n = 69 subjects could be categorized as meeting criteria for provisional MDD and n = 45 provisional PTSD based on a PCL-C cut-off of 44 or greater. For the purpose of analyses, interviewee speech content was separated from the interviewer and used for analyses while interviewer speech was omitted from analyses.

*Clinical case study assessment*

Case studies and accompanying diagnoses were taken from *DSM-5 Clinical Cases* is a companion book to the Diagnostic and Statistical Manual of Mental Disorders, Fifth Edition (DSM-5), which is the standard classification of mental disorders used by mental health professionals in the United States. The book provides in-depth case studies of patients with a variety of mental disorders, as classified by the DSM-5. The book is divided into 20 chapters, each of which corresponds to a major category of mental disorders in the DSM-5. Each chapter contains a series of case studies with distinct diagnoses within a diagnostic category. The case studies are written by leading experts in the field of psychiatry and provide detailed descriptions of the patients' symptoms, history, and treatment. The areas of psychiatry covered include Neurodevelopmental disorders, Schizophrenia spectrum and other psychotic disorders, Bipolar and related disorders, Depressive disorders, Anxiety disorders, Obsessive-compulsive and related disorders, Trauma- and stressor-related disorders, Dissociative disorders, Somatic symptom and related disorders, Feeding and eating disorders, Elimination disorders, Sleep-wake disorders, Sexual dysfunctions, Gender dysphoria, Disruptive, impulse-control, and conduct disorders, Substance-related and addictive disorders, Neurocognitive disorders

The current analyses utilized all case studies from the most common psychiatric disorder categories including Depressive disorders, Anxiety disorders, Trauma- and stressor-related disorders, substance-related and addictive disorders. And psychotic disorders.

## Outcomes

### Depression and PTSD clinical interview assessments

Primary outcome metric of PTSD severity was the PTSD Check-List Civilian Version((PCL-C ), a 17-item self-report measure of posttraumatic stress disorder (PTSD) symptoms.[17] The PCL-C is based on the DSM-IV criteria for PTSD and assesses the frequency and severity of symptoms in the past month and validated to assess non-military related symptoms of PTSD including, re-experiencing the traumatic event (e.g., nightmares, flashbacks), avoidance of reminders of the traumatic event, negative alterations in cognitions and mood (e.g., negative thoughts about oneself, difficulty feeling positive emotions), and alterations in arousal and reactivity (e.g., difficulty sleeping, irritability). Each item is rated on a scale of 1 (not at all) to 5 (extremely). The total score ranges from 17 to 85, with higher scores indicating more severe PTSD symptoms. A score of 50 or higher is considered to be indicative of PTSD. A clinical cut-off score of 44 for "probable PTSD" was selected based on guidance from the validation literature.[17]

The primary outcome metric of depression severity was the Patient Health Questionnaire 8-item version (PHQ-8).[6] The PHQ-8 is a brief, self-report questionnaire that is used to screen for and measure the severity of depressive symptoms. It is based on the nine diagnostic criteria for major depressive disorder (MDD) in the DSM-IV. The PHQ-8 is a widely used and well-validated measure of depression. The scale consists of eight items in which respondents are asked to rate how often they have experienced each symptom over the past two weeks on a scale of 0 (not at all) to 3 (nearly every day). The total score ranges from 0 to 24, with higher scores indicating more severe depressive symptoms. A cut off score of 10 or higher was selected to assign probable depression to interviews.

*Clinical case study assessment*
Outcomes accompanied each case study. Outcomes included 1) diagnostic category (e.g. depressive disorder; 2) specific disorder or diagnosis (e.g. Major Depressive Disorder) along with diagnostic modifiers or comorbidities (e.g with psychotic features).

## *Analysis*

*Depression and PTSD clinical interview assessments*
To assess the accuracy of Med-PaLM 2, we first applied prompts to focus the model's attention through the prompt: *"Are you familiar with the [PHQ-8/PCL-C]?"*. Next, Med-PaLM 2 was prompted, *"Based on the following clinical interview, what do you estimate the Participants [PHQ-8/PCL-C] score is?"* Med-PaLM 2's performance with no additional training or data examples was assessed and compared to models that were trained on example data. For each interview, Med-PaLM 2 provided: 1) an estimated clinical score; 2) a model-derived confidence score for each result and; 3) a description of the reasoning behind the score. The confidence score is a log-likelihood estimation that the model would generate the target from the given input. Based on the published literature for each scale, provisional diagnoses of major depressive disorder (MDD; PHQ-8 cutoff =>10) and posttraumatic stress disorder (PTSD; PCL-C cutoff => 44). We further compared the accuracy of model-derived diagnoses based on different cutoffs for the confidence scores. Finally, we analyzed the frequency of words and phrases taken from the MDD and PTSD diagnosis to determine if Med-PaLM 2 was more likely to use appropriate descriptors associated with each diagnosis ( e.g. PTSD: flashbacks, intrusive thoughts; MDD: low mood, loss of appetite;  see supplemental Table 3 for list of descriptive terms per diagnosis). The term "sleep" was removed because abnormal sleep is a symptom of both PTSD and MDD.

*Clinical case study assessment*
To assess the accuracy of Med-PaLM 2, each case study was entered and prompted "Take on the expertise of an expert in psychiatric diagnosis using the DSM 5. Read the following case study and apply the most appropriate diagnoses." Each case study produced a diagnosis that was used to assess both the diagnostic category and primary diagnosis. Additionally, when appropriate, Med-PaLM 2 produced secondary diagnoses and diagnostic modifiers. These outputs were used for statistical comparisons.

**Supplemental figures**

Supplemental Figure 1: Model accuracy and sample size across model-derived confidence score cut-offs.

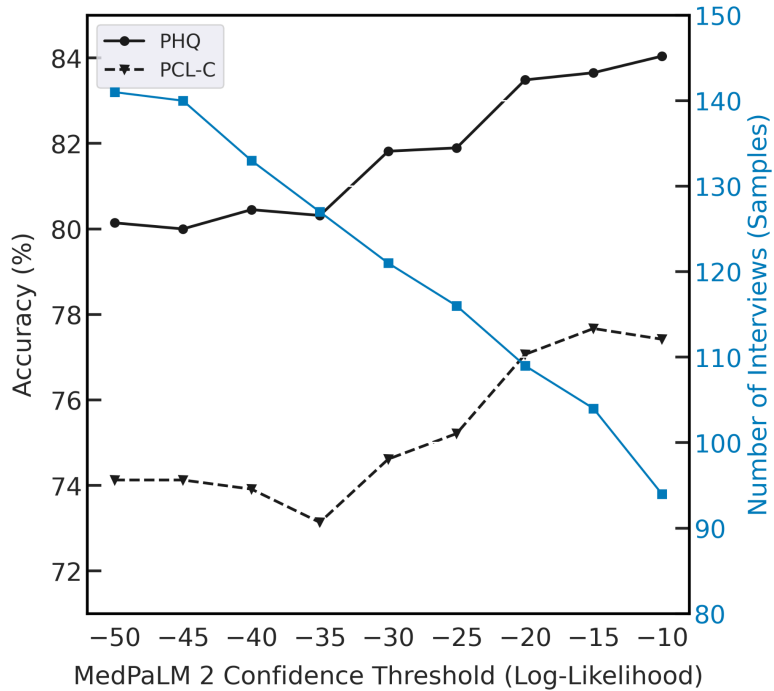

Supplemental Table 1: Frequency of words in Med-PaLM 2's reason for the associated Major Depressive Disorder (MDD) and Posttraumatic Stress Disorder (PTSD) assessment.

|           |                 | Actual   |           |       |
|-----------|-----------------|----------|-----------|-------|
|           |                 | MDD Term | PTSD Term | Total |
| Predicted | MDD Assessment  | 118      | 8         | 126   |
|           | PTSD Assessment | 26       | 101       | 127   |
|           | Total           | 144      | 109       |       |

Supplemental Table 2: Results of Med-PaLM 2 and other best in class models in prediction of PCL-C and PHQ-8 scores and clinical cutoffs.

|  | Category | Diagnosis | Fisher's Exact Test Phi(p - value) | Odds Ratio |
|---|---|---|---|---|
| Depression | 1.00 | 0.83 | -0.27( = 0.09) | 0.32 |
| Anxiety | 1.00 | 0.83 | -0.02( >0,99) | 0.09 |
| Psychosis | 0.86 | 0.71 | -0.04( >0,99) | 0.23 |
| Trauma & Stress | 0.80 | 0.60 | 0.14( = 0.58) | 0.16 |
| Addictive disorder | 1.00 | 1.00 | -0.16 ( = 0.57) | 0.25 |
| All | 0.94 | 0.71 |  |  |

*Note*: The frequency of correct diagnosis within each category was compared to the frequency of correct diagnosis across all other categories using Fisher's exact test. All significance tests are calculated based on two-tailed significance and odds ratios are calculated regarding the larger class.

Supplemental Table 3: Descriptive terms for Major Depressive Disorder and Posttraumatic Stress Disorder.

| **Depressive Disorders** | **Trauma & Stress Disorders** |
|---|---|
| *Phrases* | |
| Depressed mood | Reliving the trauma |
| Loss of interest | Negative changes in thinking and mood |
| Loss of energy | Physical reactions |
| Weight gain | Emotional reactions |
| Weight loss | Intrusive thoughts |
| Low energy | Avoid people |
| Making decisions | Avoid places |
| Change in sleep | Sleep disturbances |
| Simple tasks | Traumatic event |
| Loss of pleasures | |
| *Keywords* | |
| Appetite | Avoidance |
| Weight | Flashback |
| Sad | Nightmare |
| Hopeless | Startle |
| Empty | Hypervigilant |
| Helpless | Numb |
| Cry | Hopeless |
| Energy | Angry |
| Overeat | Concentration |
| Think | Sleep |
| Concentrate | Trauma |
| Worthless | |
| Guilt | |
| Burden | |
| Foggy | |
| Focus | |
| Attention | |
| Slow | |
| Sleep | |
| Oversleep | |
| Undersleep | |
| Task | |
| Fatigue | |
| Pleasure | |
| Agitation | |

*Note*: Keywords and phrases were taken from symptom descriptions in the Diagnostic and statistical Manual of Mental Disorders - 5th edition (DSM 5).